\documentclass[conference]{IEEEtran}
\IEEEoverridecommandlockouts
\usepackage{cite}
\usepackage{amsmath,amssymb,amsfonts}
\usepackage{algorithmic}
\usepackage{graphicx}
\usepackage{textcomp}
\usepackage{xcolor}
\usepackage{multirow}
\def\BibTeX{{\rm B\kern-.05em{\sc i\kern-.025em b}\kern-.08em
		T\kern-.1667em\lower.7ex\hbox{E}\kern-.125emX}}

\usepackage{subcaption}
\usepackage{ctable}
\usepackage{bbold}

\graphicspath{{figures/}}

\def\*#1{\mathbf{#1}}

\begin{document}
	
	\title{Channel Pruning via Multi-Criteria based on Weight Dependency}
	
	\author{\IEEEauthorblockN{Yangchun Yan\IEEEauthorrefmark{1}, Rongzuo Guo\IEEEauthorrefmark{1}, Chao Li\IEEEauthorrefmark{2}${}^{\star}$\thanks{${}^{\star}$Chao Li is the corresponding author.}, Kang Yang\IEEEauthorrefmark{2} and Yongjun Xu\IEEEauthorrefmark{2}}
		\IEEEauthorblockA{\IEEEauthorrefmark{1}College of Computer Science, Sichuan Normal University, Chengdu, China}
		\IEEEauthorblockA{\IEEEauthorrefmark{2}Institute of Computing Technology, Chinese Academy of Sciences, Beijing, China\\
			Email: \{coderyyc, gyz00001\}@163.com, \{lichao, yangkang, xyj\}@ict.ac.cn}}

	\maketitle
	
	\begin{abstract}
		Channel pruning has demonstrated its effectiveness in compressing ConvNets. In many related arts, the importance of an output feature map is only determined by its associated filter. However, these methods ignore a small part of weights in the next layer which disappears as the feature map is removed. They ignore the phenomenon of weight dependency. Besides, many pruning methods use only one criterion for evaluation and find a sweet spot of pruning structure and accuracy in a trial-and-error fashion, which can be time-consuming. In this paper, we proposed a channel pruning algorithm via multi-criteria based on weight dependency, CPMC, which can compress a pre-trained model directly. CPMC defines channel importance in three aspects, including its associated weight value, computational cost, and parameter quantity. According to the phenomenon of weight dependency, CPMC gets channel importance by assessing its associated filter and the corresponding partial weights in the next layer. Then CPMC uses global normalization to achieve cross-layer comparison. Finally, CPMC removes less important channels by global ranking. CPMC can compress various CNN models, including VGGNet, ResNet, and DenseNet on various image classification datasets. Extensive experiments have shown CPMC outperforms the others significantly. 
	\end{abstract}
	
	
	\section{Introduction}
	\label{sec:intro}
	In recent years, the growing demands of deploying CNN models to resource-constrained devices such as FPGA and mobile phones have posed great challenges. Network pruning has become one of the most effective methods to compress the model with minimal loss in performance. Network pruning can be divided into two categories: weight-level pruning and structural pruning.
	
	The weight-level pruning\cite{Han2016,Guo2016} tries to detect the redundant weights and set them to zero. It contributes little to compress deep models unless users use specialized libraries that support sparse matrix calculation. Unfortunately, the support for these libraries on resource-constrained devices like FPGA is limited. At the same time, structural pruning\cite{Li2017,Liu2017,He2019,He2018,He2017} can be a solution to this problem. These methods evaluate and remove structure weights like filters and channels in convolutional layers or unimportant nodes in fully connected layers \cite{Wang2019}. In this way, compressing the deep models is more efficient. 
	
	Channel pruning is a specific method of structural pruning \cite{He2017}, assessing the importance of output feature maps, and removing all weights which are associated with those unimportant feature maps. A feature map is considered to be a channel for output. There is no doubt that how to evaluate a channel is the key factor. We find the current state-of-the-art methods have at least one of the following issues.
	
	\textbf{Neglect of weight dependency}. Many existing criteria \cite{Li2020,Liu2017,He2018} of measuring the importance of a filter or a channel only consider the weights of a filter, and have little consideration to a part of weights that disappear with the associated filter. They ignore weight dependency.
	
	\textbf{Trial-and-error fashion}. As the redundancy of each layer in a deep model is various, a different number of filters or channels should be pruned in each layer. In the meanwhile, some pruning methods adopt intra-layer comparison. Therefore, they require users to specify the layerwise pruning ratios manually or automatically\cite{Guo2020,He2017,Lin2020}. They all use trial and error fashion to get the layerwise pruning ratios, which is less efficient. 
	
	\textbf{No multi-criteria}. Many methods use only one criterion for evaluation\cite{Luo2017,HRank,Molchanov2019}. At different positions, pruning a channel can reduce the different number of parameters or FLOPs, or both. computational cost and parameter quantity are essential for compressing models. They do not add them to the criteria.
	
\begin{figure*}[ttt]
		\centering
		\includegraphics[width=160mm,clip]{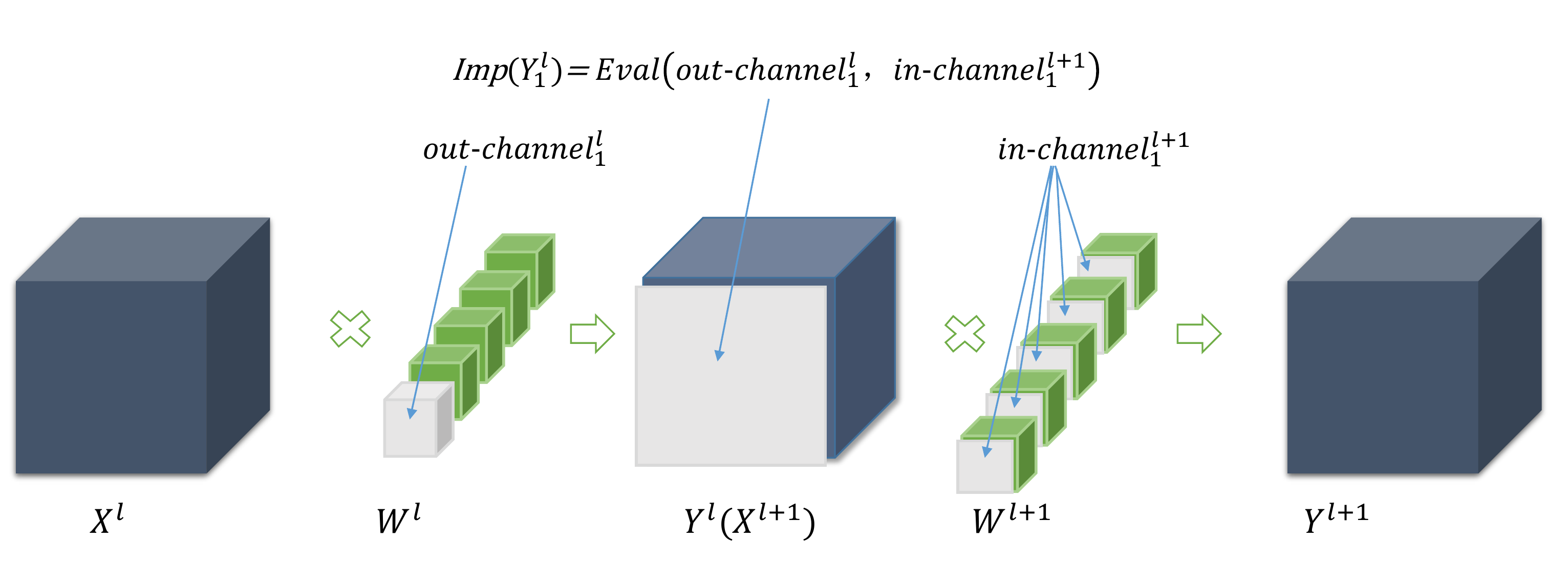}  
		\vspace{-5pt}
		\caption{Weight dependency between two consecutive layers and the definitions of out-channel and in-channel in a common model. \emph{${out-channel}_1^l$} is the associated filter of the channel \emph{$Y_1^l$}. \emph{${in-channel}_1^l$} consists of several corresponding convolutional kernels in the next layer. In the $l_{th}$ layer, \emph{$X^l$} is the input, \emph{$W^l$} is the weight matrix, \emph{$Y^l$} is the output and it is convoluted by out-channels. Next, \emph{$Y^l$} is treated as the input of the next layer. When we prune the channel \emph{$Y_1^l$}, its out-channel and in-channel are also removed. Therefore, \emph{${out-channel}_1^l$} and \emph{${in-channel}_1^l$} should be evaluated together.
		}
		\label{fig:dependency}
		\vspace{-2pt}
	\end{figure*}

	To address the above issues, we develop a channel pruning method via multi-criteria based on weight dependency(CPMC). As shown in Figure \ref{fig:dependency}, we define the importance of a channel by its associated filter, called out-channel, and some structural weights in the next layer, called in-channel. In addition, we use multi-criteria to evaluate the importance of the channel, including its associated weight value, computational cost, and parameter quantity. For every criterion, we adopt appropriate normalization methods. Finally, we globally rank the channel importance of different layers, which can avoid the trial-and-error fashion. Our method can compress directly a pre-trained model, which enables the users to customize the compression according to preference more efficiently.
	
	It is worthy to highlight the advantages of CPMC:
	\begin{enumerate}
		\item We propose a novel channel-level pruning method for deep model compression and acceleration, which can (1) remove unimportant channels significantly, (2)  evaluate the importance of a channel by multi-criteria, and (3) enable compress directly a pre-trained model and avoid the trial-and-error fashion.
		\item Extensive experiments on public datasets have demonstrated CPMC’s advantages over some current methods.
		\item We also evaluate CPMC on a specially designed CNN model, like DenseNet. CPMC still produces reasonably good compression and acceleration ratio with little loss.
	\end{enumerate}
	\section{Related Work}
	\label{sec:RW}
	
	Compacting CNN models for speeding up inference and reducing storage overhead has been an influential project in both academia and industry.
	
	Recently, much attention has been focused on structural pruning methods to reduce model parameters and FLOPs. Li \emph{et al.}\cite{Li2017} evaluated the importance of filters through its $l_1 norm$. He \emph{et al.}\cite{He2018} proposed a soft filter pruning method that can let the pruned filters be updated in the training stage. He \emph{et al.}\cite{He2019} compressed models by pruning filters with the most replaceable contribution which calculated by the Geometric Median. He \emph{et al.}\cite{He2017} evaluated the importance of an output feature map by a LASSO regression based channel selection and least square reconstruction. Lin \emph{et al.}\cite{HRank} proposed a filter pruning method by exploring the high rank of feature maps and they believe that low-rank feature maps contain less information. These methods provided some effective criteria for pruning models, but they all required users to set pruned ratio for each layer manually. Lately, Liu \emph{et al.}\cite{Liu2019} applied an evolutionary algorithm to get the layerwise pruning ratios automatically. He \emph{et al.}\cite{AMC}  proposed a pruning framework which set an optimized pruning ratio for each layer based on searching via reinforcement learning. All the methods mentioned above defined the importance locally within each layer which could only be compared within each layer. Therefore, they needed a pruning ratio combination for different convolutional layers. 
	
	More recent developments adopted global comparison to avoid the layerwise pruning ratios. Liu \emph{et al.}\cite{Liu2017} added a sparsity regularization into loss function and used the scale of batch normalization layer as the global importance. Wang \emph{et al.}\cite{Wang2019} developed a filter level algorithm which evaluated the importance of filters by Pearson correlation. Meanwhile, they globally ranked the importance and added layerwise regularization terms to improve the effect. Chin \emph{et al.}\cite{Chin2020} proposed the learned global ranking which used the regularized evolutionary algorithm to produce a set of pruned CNN models with different performances. All the methods mentioned above had little consideration to the weight dependency. They only evaluated the importance of out-channel but in-channels are neglected.
	
	Recently, fewer pruning methods had been focused on the phenomenon of weight dependency. Li \emph{et al.}\cite{oicsr}concatenated out-channels and in-channels as one regularization and added the structural sparsity regularization into loss function. And they used Group Lasso to define the importance of channels. To get better results, they needed to prune iteratively. Despite their success, we noticed that they failed to prune the pre-trained models due to regularization and have no consideration of multiple criteria, especially parameter quantity and computational cost.
	
	\begin{figure}[ttt]
		\centering
		\includegraphics[width=85mm,clip]{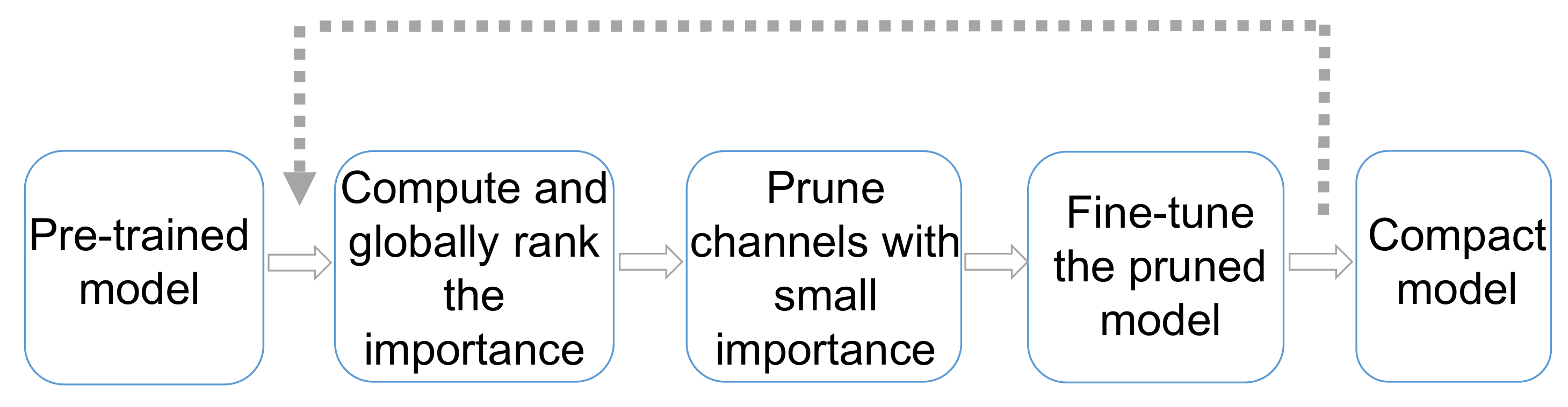}  
		\vspace{-5pt}
		\caption{The overview of the proposed channel pruning. The dotted line is for an optional iterative scheme.}
		\label{fig:stages}
		\vspace{-2pt}
	\end{figure}
	
	To improve the efficiency of the model in the inference stage, some works explored skipping part of the model based on each input and proposed dynamic pruning methods\cite{Gao2019,Zoph2018}. Unlike static pruning methods which result in a fixed pruned model for all inputs, dynamic pruning methods dynamically choose the part of the model to inference for each input. For example, Wu \emph{et al.}\cite{Wu2018} proposed an approach that learns to dynamically choose layers during inference. Dynamic pruning methods succeed to achieve the better acceleration of models due to instance-wise sparsity, but they tend to make the actual inference speed slower because of the computational cost of reindexing the dynamic model structure for each input\cite{instance-wise}. This paper is centered on static pruning methods.  
	
	Low rank approximation, knowledge distillation, network quantization, and the lightweight model design are also popular techniques to obtain more compact models. (1) Low rank approximation reduces computation by decomposing large matrices into some small matrices\cite{Peng2018,Wen2017}. (2) Knowledge distillation lets a small student model get the learned knowledge from one or more large teacher models\cite{Frosst2018,Fukuda2017,yang2020multi}. (3) Network quantization quantizes the weights into fewer bits to reduce model complexity\cite{Rastegari2016,Yin2019}. (4) The lightweight model design aims to more compact CNN architectures. For example, SqueezeNet\cite{Gholami2018}, MobileNet\cite{Howard2017}, HCGNet\cite{yang2020gated}. Combing with our channel pruning, these techniques have further improvement.

	\section{ALGORITHM}
	\label{sec:algorithm}
	
	\begin{figure*}[ttt]
		\centering
		\includegraphics[width=160mm,clip]{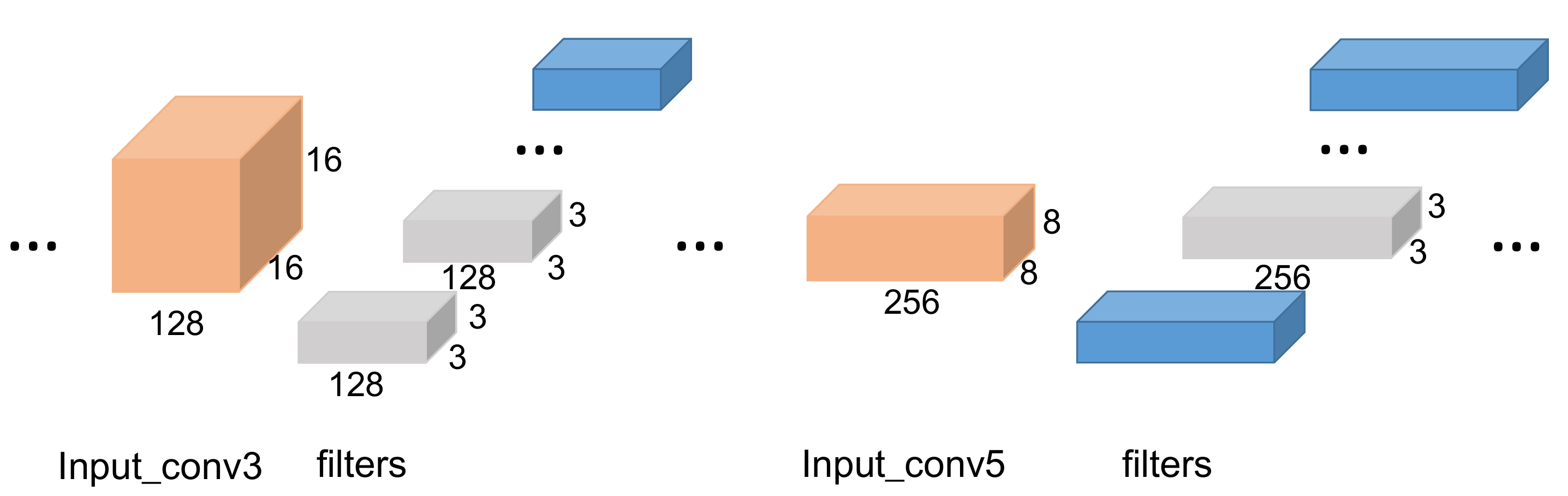}  
		\vspace{-5pt}
		\caption{The figure is a simple sketch of VGG16 on CIFAR10 dataset. In order to reduce the same number of parameters, we either prune two out-channels in the $3_{th}$ layer, or prune one out-channel in the $5_{th}$ layer. It is worth noting that the two pruning strategies mentioned above produce different acceleration effects. In terms of reduced computation cost, the latter is a quarter of the former. Therefore, an out-channel in different layers brings a different number of parameters or FLOPs. And the same thing is true for in-channels. To get more compact models, we make CPMC aware of such differences.}
		\label{fig:cost}
		\vspace{-2pt}
	\end{figure*}
	
	\subsection{Overview}
	\label{sec:overview}
	
	We aim to provide a simple and efficient channel pruning framework to achieve channel level pruning in deep CNNs. Our channel pruning procedures are illustrated in Figure \ref{fig:stages}. Particularly, we start with a pre-trained model and have \emph{NO} sparsity regularization during the pre-trained stage, which enables the users to compress models more efficiently. when we get the local importance of each channel in the whole model, it is puzzling to determine the layerwise channel pruning ratios. Therefore, we use global normalization to achieve cross-layer comparison within a whole model, instead of comparison within a layer. 
	
	Moreover, when choosing the unimportant channels, we sort the final importances and get the threshold value such that the FLOP count is met. Then, we prune the channels whose final importances are below the threshold. 
	
	Finally, we fine-tune the pruned model to recover the accuracy. We can also extend the proposed method from a single-pass pruning scheme to an iterative multi-pass one and prune smoothly in each pruning iteration to get a more compact model.
	
	\subsection{Criteria of channel importance}
	\label{sec:criterion}
	There is no doubt that the criteria of a channel are the keys. We take a simple model as an example and introduce the calculation method for the final importance of a channel. Simply, in a deep model which has \emph{L} layers and produces \emph{S} output feature maps in total, the input of the $l_{th}$ layer is \emph{$X^l$}, the output is \emph{$Y^l$} and the weight matrix is \emph{$W^l$}. \emph{$Y_i^l$} is a channel or a node for output.
	
	When the $l_{th}$ layer in the deep model is a convolutional layer, \emph{$W^l$} is of shape \emph{$M^l$} $\times$ \emph{$N^l$} $\times$ \emph{$K^l$} $\times$ \emph{$K^l$}, \emph{$M^l$} is the number of input channel, \emph{$N^l$} is the number of output channel, \emph{$K^l$} $\times$ \emph{$K^l$} is the size of convolutional kernels. \emph{$X^l$} is of shape \emph{$I^l$} $\times$ \emph{$I^l$} $\times$ \emph{$M^l$}, \emph{$I^l$} $\times$ \emph{$I^l$} is the size of input feature maps of the $l_{th}$ layer. \emph{$Y^l$} is of shape \emph{$O^l$} $\times$ \emph{$O^l$} $\times$ \emph{$N^l$}. \emph{$O^l$} $\times$ \emph{$O^l$} is the size of output feature maps of the $l_{th}$ layer.
	
	When the $l_{th}$ layer in the deep model is a fully connected layer, \emph{$W^l$} is of shape \emph{$M^l$} $\times$ \emph{$N^l$}, its shape be treated as \emph{$M^l$} $\times$ \emph{$N^l$} $\times$ 1 $\times$ 1, which makes that the calculation methods become consistent. \emph{$M^l$} is the number of input nodes and \emph{$N^l$} is the number of output nodes. \emph{$X^l$} is of shape \emph{$M^l$} $\times$ 1, \emph{$Y^l$} is of shape \emph{$N^l$} $\times$ 1.
	
	\textbf{Weight dependency}. Channel pruning is designed to reduce the number of output feature maps\cite{He2017} and then prune the weights of some filters and corresponding kernels of each filter in the next layer. We start by analyzing the prior methods and find that many methods only consider out-channel to represent the importance of an output feature map. They have little consideration for the importance of in-channel. When pruning an output feature map whose in-channel is important, some avoidable performance losses result. Therefore, we evaluate out-channel and in-channel together and define the importance of an output feature map as: 
	\begin{align}
		\label{equ:imp}
		Imp(Y_{i}^{l}) = Eval({OC}_i^l , {IC}_i^{l+1})
	\end{align}
	Where \emph{$Imp(Y_{i}^{l})$} is the importance of \emph{$Y_{i}^{l}$} of the $l_{th}$ layer, \emph{${OC}_i^l $} is the out-channel which is the associated filter in current layer. \emph{${IC}_i^{l+1}$} is the in-channel which is some corresponding kernels of each filter in the next layer. Respectively, \emph{$Eval({OC}_i^l,{IC}_i^{l+1})$} is the evaluation value by out-channel and in-channel, more details about it are given in the following part.
	
	\textbf{Multi-criteria}. Many methods do not evaluate structural weights from multiple perspectives, especially in terms of computational cost and parameter quantity. Our multi-criteria consists of three parts, including weight value, computational cost, and parameter quantity. 
	
	We note that the \emph{norm assumption} is adopted and empirically verified by prior art \cite{Li2017,He2018}. In the paper \cite{Li2017}, they propose to measure the relative importance of a filter in each layer by using \emph{$l_1 norm$}, it is defined as follows:
	\begin{align}
		L_{i}^{l} = \left \| W_{i,:}^{l} \right \|_{1} = \sum_{j}\left | W_{i,j}^{l} \right | 
	\end{align} 
	where $L_{i}^{l}$ is just only the evaluation value in weight value of the out-channel of $Y_{i}^{l}$, \emph{$W^l_{i,:}$} means the weight matrix of the $i_{th}$ filter in the $l_{th}$ layer, \emph{$W^l_{i,j}$} is the weight matrix of the $j_{th}$ convolutional kernel of the $i_{th}$ filter. The importance of in-channel is ignored, which may produce several incorrect selections of redundant channels. Therefore, we measure the importance of a channel by evaluating its out-channel and in-channel together. The evaluation value based on weight dependency is given as:
	\begin{align}
		L_{i}^{l,l+1} = \left \| W_{i,:}^{l} \oplus W_{:,i}^{l+1} \right \|_{1} = \sum_{j}\left | W_{i,j}^{l} \right | + \sum_{j}\left | W_{j,i}^{l+1} \right |
	\end{align} 
	where $L_{i}^{l,l+1}$ is the evaluation value calculated by out-channel and in-channel of $Y_{i}^{l}$ in weight value aspect, \emph{$W^l_{i,:}$} and \emph{$W^{l+1}_{:,i}$} mean the weight matrices of the out-channel and in-channel of $Y_{i}^{l}$ respectively.
	
	However, \emph{$l_1 norm$} can be used within each layer, but not across layers. Due to different functions and scopes, the weight value in the different layers may not be in the same order of magnitude. For cross-layer comparison, we need to normalize the evaluation value. After analysis and experiment, we propose to use \emph{max-min} normalization which is a linear transformation. We normalize the correlation distribution of each layer to align the correlation distribution to [0,1]. Formally, we define the normalized evaluation value as:
	\begin{align}
		GL_{i}^{l,l+1} &= \frac{L_{i}^{l,l+1}-L_{p}^{l,l+1}}{L_{q}^{l,l+1}-L_{p}^{l,l+1}}\nonumber \\ L_{p}^{l,l+1} &= \min_i L_{i}^{l,l+1},\nonumber \\
		L_{q}^{l,l+1} &= \max_i L_{i}^{l,l+1}, p,q \in [1,N^l] \ and \ p \neq q
	\end{align} 	
	where $L_{p}^{l,l+1}$ and $L_{q}^{l,l+1}$ are the minimum and maximum evaluation value in weight value aspect of $Y^{l}$ which is the output of the $l_{th}$ layer.

	\begin{figure}[ttt]
		\centering
		\includegraphics[width=80mm,clip]{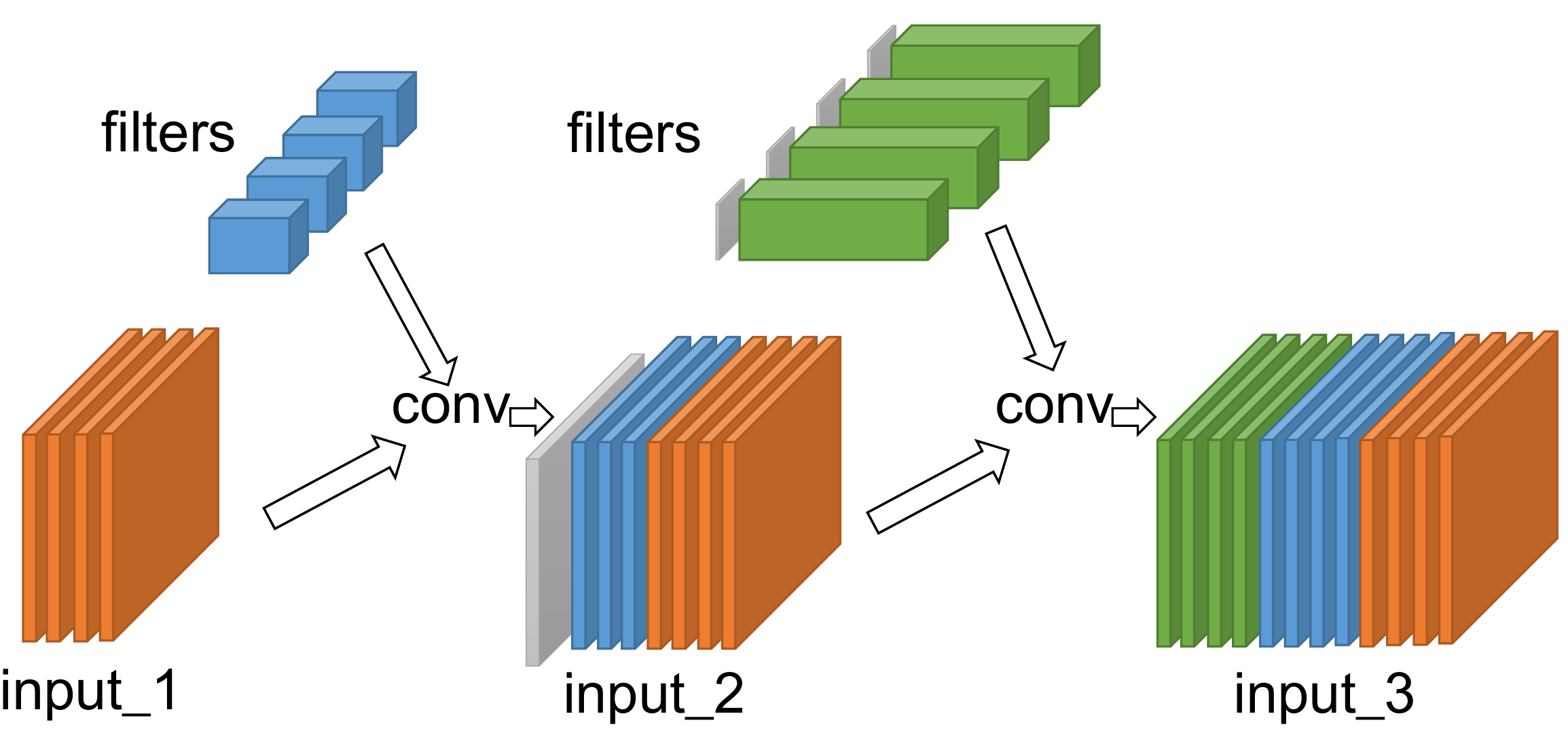}  
		\vspace{-5pt}
		\caption{The pruning strategy of a dense block. The same channels are marked by the same color. In a dense block, the output feature maps of all preceding layers are used as inputs for each layer. Therefore, input\_2 is the part of input of many layers. When pruning a channel of input\_2, CPMC only remove its in-channel and save its out-channel, because its out-channel is still evaluated in subsequent layers.}
		\label{fig:dense}
		\vspace{-2pt}
	\end{figure}
	
	\begin{table*}[ttt]
		\vspace{-10pt}
		\centering
		\caption{Pruning results on CIFAR10.}
		\label{tbl:cifar10} 
		\setlength{\tabcolsep}{4.0mm}
		\footnotesize
		\begin{tabular}{c|cccccc} \hline
			Model &           Alg 		&	 Acc(\%)		&	Param    & Prr(\%)   	 	&	FLOPs    &Frr(\%)		    \\ \hline
			
			\multirow{5}{*}{VGG-16}
			&{\tt Baseline}					&	$93.68$			&	$ 14.72M$	   & $-$	&	$ 314.15M$	& $-$	\\
			&{\tt GAL-0.05} \cite{GAL}		&	$92.03$			&	$ 3.36M$	   &$77.6$	&	$ 189.49M$ & $39.6$ 	\\ 
			&{\tt VCNNP} \cite{VCNNP}		& 	$93.18$	  		&	$ 3.92M$ & $73.3$		& 	$ 190.01M$ & $39.1$	   \\
			&{\tt HRank} \cite{HRank}		&	$92.34$			&	$ 2.64M$ & $82.1$		&	$ 108.61M$ & $65.3$	\\
			&{\tt CPMC(Ours)}  				&	${\bf93.40}$	&	$ 1.04M$ & ${\bf 92.9}$	&	$ 106.68M$ & ${\bf 66.0}$	\\ 	
			\hline	
			\multirow{3}{*}{ResNet-20}
			&{\tt Baseline}			 		&	$91.30$			&	$220.39K$	& $-$	&	$34.62M$	& $-$\\ 
			&{\tt NS}\cite{Liu2017}			&	$90.04$			&	$189.70K$	& $13.9$	&	$27.61M$ & $20.2$		\\ 
			&{\tt CPMC(Ours)}				&	${\bf 91.13}$	&	$147.88K$ & ${\bf32.9 }$	&$24.42M$ & ${\bf 29.5}$\\ 
			\hline
			\multirow{3}{*}{ResNet-56}
			&{\tt Baseline}			 		&	$93.72$			&	$593.13K$ & $-$		&	$89.59M$	& $-$\\ 
			&{\tt NS}\cite{Liu2017}			&	$92.84$			&	$441.59K$	& $25.5$	&	$59.91M$ & ${\bf 33.1}$	\\ 
			&{\tt CPMC(Ours)}				&	${\bf 93.46}$	&	$341.10K$ & ${\bf 42.5}$	&	$60.43M$ & $32.5$         \\
			\hline
			\multirow{3}{*}{ResNet-164}
			&{\tt Baseline}			 		&	$95.01$			&	$1.71M$	& $-$	&	$254.50M$ & $-$	\\ 
			&{\tt NS}\cite{Liu2017}			&	$94.73$			&	$1.10M$ & $35.2$		&	$137.50M$ & ${\bf 44.9}$\\ 
			&{\tt CPMC(Ours)}				&	${\bf 94.76}$	&	$0.75M$ & ${\bf 56.0}$	&	$144.02M$ & $43.4$      \\
			\hline
			\multirow{5}{*}{DenseNet-40}
			&{\tt Baseline}			 		&	$94.21$			&	$1.06M$	& $-$		&	$290.13M$ & $-$	\\ 
			&{\tt GAL-0.01}\cite{GAL}		&	${\bf 94.29}$	&	$0.67M$ & $35.6$		&	$182.92M$ & $35.3$\\
			&{\tt HRank} \cite{HRank}		&	$94.24$			&	$0.66M$ & $36.5$		&	$167.41M$ & $40.8$	\\ 
			&{\tt VCNNP} \cite{VCNNP}		& 	$93.16$			&	$0.42M$ & $59.7$		& 	$ 156.00M$ & $44.8$	  \\
			&{\tt CPMC(Ours)}				&	$93.74$			&	$0.42M$ & ${\bf 60.7}$	&	$121.73M$ & ${\bf 58.0}$ \\
			\hline
		\end{tabular}
		
	\end{table*}

	\begin{table*}[htt]
		\vspace{-10pt}
		\centering
		\caption{Pruning results on CIFAR100.} 
		\label{tbl:cifar100}
		\setlength{\tabcolsep}{4.0mm}
		\footnotesize
		\begin{tabular}{c|cccccc} \hline
			Model &           Alg 		&	 Acc(\%)		&	Param & Prr(\%)  	 	&	FLOPs	& Frr(\%)	    \\ \hline
			
			\multirow{4}{*}{VGG-16}
			&{\tt Baseline}					&	$73.80$			&	$ 14.77M$ & $-$		&	$ 314.2M$ & $-$			\\
			&{\tt VCNNP} \cite{VCNNP}		& 	$73.33$   		&	$ 9.14M$ & $37.9$		& 	$ 256.00M$ & $18.0$	    \\
			&{\tt CPGMI} \cite{CPGMI}		&	${\bf73.53}$	&	$ 4.99M$ & $66.8$		&	$ 198.20M$ & $37.1$  	\\ 
			&{\tt CPMC(Ours)}  				&	$73.01$			&	$ 4.80M$ & ${\bf 67.5}$	&	$ 162.00M$ & ${\bf 48.4}$	\\ 	
			\hline	
			\multirow{3}{*}{ResNet-56}
			&{\tt Baseline}			 		&	$73.84$			&	$616.26K$ & $-$		&	$89.61M$ & $-$	\\ 
			&{\tt NS}\cite{Liu2017}			&	${\bf 73.36}$	&	$550.26K$ & $18.8$		&	$60.55M$ & $32.4$	\\ 
			&{\tt CPMC(Ours)}				&	$73.31$	 		&	$298.57K$ & ${\bf 51.6}$	&	$49.02M$ & ${\bf 45.3}$       \\
			\hline
			\multirow{3}{*}{ResNet-164}
			&{\tt Baseline}			 		&	$77.00$			&	$1.73M$ & $-$		&	$254.52M$ & $-$	\\ 
			&{\tt NS}\cite{Liu2017}			&	$76.18$			&	$1.21M$ & $29.7$		&	$123.50M$ & ${\bf 50.6}$	\\ 
			&{\tt CPMC(Ours)}				&	${\bf 77.22}$	&	$0.96M$ & ${\bf 44.9}$	&	$151.92M$ & $40.3$          \\
			\hline
			\multirow{4}{*}{DenseNet-40}
			&{\tt Baseline}					&	$74.31$			&	$ 1.11M$ & $-$			&	$ 290.18M$ & $-$			\\
			&{\tt VCNNP} \cite{VCNNP}		& 	$72.19$   		&	$ 0.65M$ & $37.5$		& 	$ 218.00M$ & $22.5$	    \\
			&{\tt CPGMI} \cite{CPGMI}		&	$73.84$			&	$ 0.66M$ & $40.5$		&	$ 198.50M$ & $32.0$  	\\ 
			&{\tt CPMC(Ours)}  				&	${\bf 73.93}$	&	$ 0.58M$ & ${\bf 47.4}$	&	$ 155.24M$ & ${\bf 46.5}$	\\ 
			\hline
		\end{tabular}
	\end{table*}
	As shown in Figure \ref{fig:cost}, pruning a channel in different layers can reduce the different number of parameters or FLOPs or both. To make our approach aware of such differences, we add two criteria to better model compression. Specifically, we add the evaluations of parameter quantity and computational cost. Pruning channel $Y_{i}^{l}$ can reduce the parameter quantity and computational cost of its out-channel and in-channel. So the parameter quantity and computational cost related with channel $Y_{i}^{l}$ are defined as:
	\begin{align}
		P_{i}^{l,l+1} &= K^lK^lM^l + K^{l+1}K^{l+1}N^{l+1} \nonumber \\
		F_{i}^{l,l+1} &= 2I^lI^lK^lK^lM^l + 2I^{l+1}I^{l+1}K^{l+1}K^{l+1}N^{l+1}
	\end{align} 
	where $P_{i}^{l,l+1}$, $F_{i}^{l,l+1}$ are the parameter quantity and computational cost respectively.
	On the premise of no obvious decline of inference ability, we wish to minimize the resource consumption of the model on the device. So parameter quantity and computational cost are treated as the two criteria to evaluate the importance of channel $Y_{i}^{l}$.
	
	In the same way, we need to normalize parameter quantity and computational cost to align their values to [0,1]. We propose to use $log$-normalization because their values are large. Formally, in order to get more compact models, we define the normalized evaluation values of parameter quantity and computational cost as: 
	\begin{align}
		GP_{i}^{l,l+1} &= \alpha (1-\frac{log(P_{i}^{l,l+1})}{log(P_{m}^{l,l+1})}) \nonumber \\
		GF_{i}^{l,l+1} &= \beta (1-\frac{log(F_{i}^{l,l+1})}{log(F_{n}^{l,l+1})}) \nonumber \\
		P_{m}^{l,l+1} &= \max_i P_{i}^{l,l+1}, \nonumber \\
		F_{n}^{l,l+1} &= \max_i F_{i}^{l,l+1}, m,n \in [1,S] 
	\end{align}
	where $GP_{i}^{l,l+1}$ and $GF_{i}^{l,l+1}$ are the evaluation values of parameter quantity and computational cost respectively. $P_{m}^{l,l+1}$ and $F_{n}^{l,l+1}$ are the maximum evaluation values of parameter quantity and computational cost respectively in all convolutional layers. $\alpha$ and $\beta$ are set to adapt to the differences of different models.
	
	Finally, we combine Equation 1 and define the importance of $Y_i^l$ specifically as:
	\begin{align}
		Imp(Y_{i}^{l}) = GL_{i}^{l,l+1} + GP_{i}^{l,l+1} + GF_{i}^{l,l+1}
	\end{align}
	where $Imp(Y_{i}^{l})$ is the complete importance value of channel $Y_{i}^{l}$ and is calculated by multi-criteria based on weight dependency.

	Our CPMC algorithm is a three-step pipeline: \textbf{1)} Calculate the importance value for all channels and nodes; \textbf{2)} Sort these values and remove the least important ones according to the pre-set model pruning ratio. \textbf{3)} Fine-tune the pruned model with original data. 
	
	\textbf{Pruning for residual block}. A residual block contains a cross-layer connection and more than one convolutional layer. The number of input and output feature maps must be equal for a residual block unless the shortcuts go across feature maps of two sizes \cite{He2016}. Therefore for all residual blocks whose number of input and output feature maps must be equal, we compute the mean as the importance of input and output feature maps and prune them simultaneously, which enable input and output feature maps to be aligned.
	
	\textbf{Pruning for dense block}. In a dense block, many input feature maps of a layer are from preceding layers \cite{Huang2017}. Similarly, their out-channels are scattered in preceding layers. Please refer to Figure \ref{fig:dense} for details. We take a dense block whose growth rate of $k$=4 as an example. All the blue feature maps in different layers have common out-channels. When computing the importance, we still can consider its out-channel and in-channel to represent the importance of an output feature map. But pruning channels in the current layer, we can only remove their in-channels.

	\section{Experiments and Results}
	\label{sec:ER}
	\subsection{Experimental settings}
	\label{sec:setting}
	We empirically demonstrate the effectiveness of CPMC on public datasets such as CIFAR10/100\cite{Krizhevsky2009} and SVHN datasets\cite{Netzer2011}. CIFAR10/100 and SVHN datasets are widely used in related research because they provide a quick way to verify that the method is effective \cite{CPGMI,Pollot2020}. CIFAR10/100 datasets consist of natural images of size $32\times32$ pixels. CIFAR10 and CIFAR100 are drawn from 10 classes and 100 classes respectively. The two datasets contain 50,000 train images and 10,000 test images. SVHN is the street view house number dataset and consists of $32\times32$ colored digit images. The train and test sets contain 73,257 digits and 26,032 digits respectively. A standard data augmentation scheme\cite{Liu2017,recognition,Huang2016} is adopted.

	We test the compression performance of different methods on several famous large CNN models, including VGGNet \cite{Simonyan2015}, ResNet \cite{He2016}, and DenseNet \cite{Huang2017}. Note that, we implement the compression of ResNet with “bottleneck” which is more economical and has lower complexity, especially in deeper nets \cite{He2016}. This architecture is a stack of three layers including $1\times1$, $3\times3$, and $1\times1$ convolutions instead of two $3\times3$ convolution layers. 
	
	Following the previous works \cite{Liu2017,Wang2019}, we record the parameter-reduction ratio(Prr), FLOPs-reduction ratio(Frr), and accuracy of each algorithm compared with the original model. A higher parameter-reduction ratio means the more compact pruned model. A higher FLOPs-reduction ratio means the faster pruned model.
	
	We set the batch-size of SGD to be 64 for 160 epochs. The initial learning rate is set to 0.1 and is divided by 10 at 50\% and 75\% of the total number of training epochs. We use a weight decay of $10^{-4}$ and set the momentum coefficient to be 0.9 for all models. the baseline results and the experimental results follow the same training strategies.
	
	We set $\alpha=3,\beta=1$ for VGGNet, $\alpha=1,\beta=1$ for ResNet, and $\alpha=0.1,\beta=0.1$ for DenseNet to let our method adapt to the differences of model structure.
	
	\begin{figure}[ttt]
		\centering
		\includegraphics[width=90mm,clip]{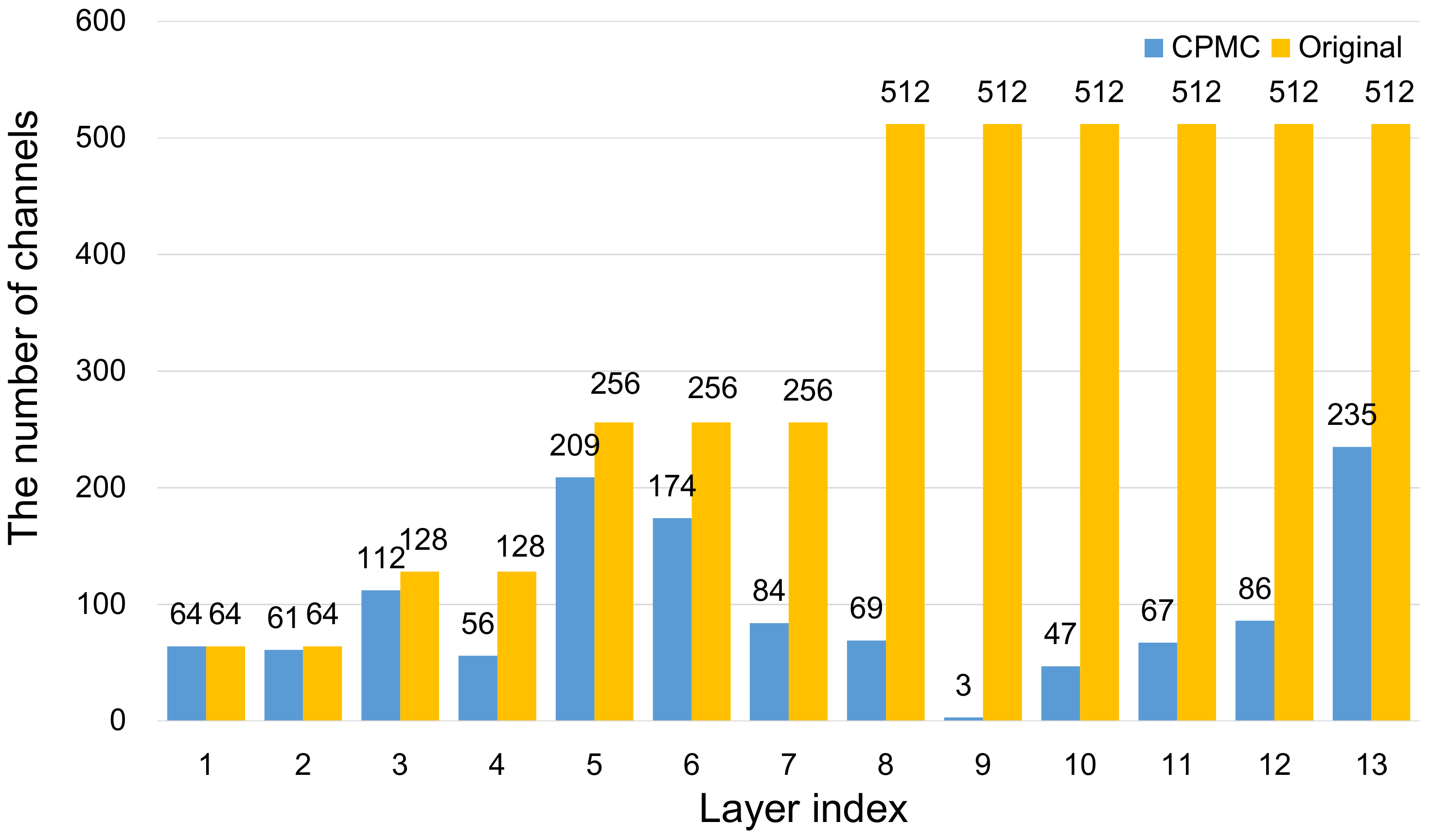}  
		\vspace{-5pt}
		\caption{The number of channels before and after pruning for VGG-16 on CIFAR10. We can see that the pruned channels are concentrated in the middle of the model}
		\label{fig:layer}
		\vspace{-2pt}
	\end{figure}
	\subsection{Results and analysis}
	\label{sec:results}
	We choose several recent state-of-the-art methods which are all static structural pruning algorithms. (1) GAL\cite{GAL}is an effective end-to-end pruning approach that uses generative adversarial learning (GAL) to solve the structure optimization problem. (2) VCNNP\cite{VCNNP} is a variational Bayesian framework for channel pruning. (3) HRank\cite{HRank} is a novel pruning method that treats the High Rank of feature maps as the importance of the filters. (4) NS\cite{Li2017} is an efficient pruning method that uses the BN’s scaling factors to evaluate the channels. (5) CPGMI\cite{CPGMI} is a newer method that uses gradients of mutual information to measure the importance of channels.
	
	We analyze the performance on CIFAR10/100, comparing against several popular CNNs, including VGG-16, ResNet-20/56/164, and DenseNet-40. We experiment with several pruned ratios on CPMC and choose the maximum pruned ratio under the constraint of acceptable accuracy loss. Then, we experiment many times and record the mean.
	
	\textbf{Results on CIFAR10}. The performance of different algorithms is given in Table \ref{tbl:cifar10}. In most experiments, CPMC can achieve a higher compression ratio and speedup ratio with higher accuracy than other algorithms. Although GAL gets higher accuracy than CPMC for DenseNet-40 (94.29\% vs. 93.74\%), CPMC provides significantly better parameters and FLOPs reductions than GAL (60.7\% vs. 35.6\% and 58.0\% vs.35.3\%).  
	
	\textbf{Results on CIFAR100}. We summarize the results on CIFAR100 in Table \ref{tbl:cifar100}. With similar accuracy, CPMC surpasses its counterparts in the parameters and FLOPs reductions for VGG-16, ResNet-56, and DenseNet-40. For ResNet-164, CPMC achieves better accuracy than Baseline (77.22\% vs. 77.00\%) and provides remarkable parameters and FLOPs reductions.
	
	As is shown in Table \ref{tbl:cifar10} and Table \ref{tbl:cifar100}, we can draw some conclusions as follows:
	\begin{enumerate}
		\item In most experiments, higher compression ratio and speedup ratio with similar accuracy to other algorithms are realized by CPMC. Especially, CPMC outperforms in light-weight ResNet. This is the contribution of the multi-criteria of CPMC which evaluate channels in parameter quantity and computational cost.
		\item For the algorithms which can prune a similar number of parameters or computational cost, CPMC gets a higher accuracy(\emph{e.g.}, VCNNP, HRank). Because CPMC considers the weight dependency, which enables many important weights of in-channel to be retained.
		\item CPMC is an interesting method that globally ranks the importance values. Compared with some methods which define the importance locally (\emph{e.g.}, HRank), it is not only user-friendly but also more efficient. In the comparison between CPMC and other global pruning methods (\emph{e.g.}, NS), CPMC still has good performance.
		
	\end{enumerate}
	
	\begin{table}[ttt] 
		\vspace{-10pt}
		\centering
		\caption{Pruning results of using different normalization.} 
		\label{tbl:norm}
		\setlength{\tabcolsep}{1.4mm}
		\footnotesize
		\begin{tabular}{c|lccccc} \hline
			Model &         Algorithm 		&	 Acc(\%)		&	Prr(\%)  		&	Frr(\%)	\\ \hline		
			\multirow{3}{*}{VGG-16}
			&$CPMC_{log}$					&	$93.22$			&	$ 83.6 $		&	$ 38.3$			\\
			&$CPMC_{max}$					& 	$93.32$   		&	$ 89.7 $		& 	$ 46.4$	    \\
			&$CPMC_{max-min}$				&	${\bf 93.40}$	&	$ {\bf 92.9} $	&	$ {\bf 66.0}$  	\\ 	
			\hline	
			\multirow{3}{*}{ResNet-20}
			&$CPMC_{log}$					&	$88.81$			&	$ 21.1 $		&	$ 19.4$			\\
			&$CPMC_{max}$					& 	$90.08$   		&	$ 25.6 $		& 	$ 25.5$	    \\
			&$CPMC_{max-min}$				&	${\bf 91.13}$	&	$ {\bf 32.9} $	&	$ {\bf 29.5}$  	\\ 	
			\hline
		\end{tabular}
		
	\end{table}
	
	\subsection{Ablation study}
	\label{ablation}
	\textbf{Normalization}. We normalize the value to cross-layer comparison. In weight value evaluation, we use the \emph{max-min} normalization. Before this, we have tried \emph{$log$}-normalization, \emph{max}-normalization and \emph{max-min} normalization when other settings remain unchanged. The results on CIFAR10 are shown in Table \ref{tbl:norm}. We can see that \emph{max-min} normalization can prune more parameters and FLOPs with higher accuracy, which is the best configuration.

	\textbf{Effect of pruning}. The numbers of channels per layer before and after pruning in VGG-16 on CIFAR10 are shown in Figure \ref{fig:layer}. CPMC retained 1267 (out of 4224) channels after pruning. Interestingly, we can see that fewer channels in some middle layers are retained. Exactly, they all have lower evaluation values in parameter quantity and computational cost. So this suggests that CPMC can identify the channels that consume too much in both calculations and parameters. On the contrary, such as most channels in the front layers are retained, because they just have a few associated parameters even though they have many computational costs.

\begin{figure}[ttt]
		\centering
		\includegraphics[width=80mm,clip]{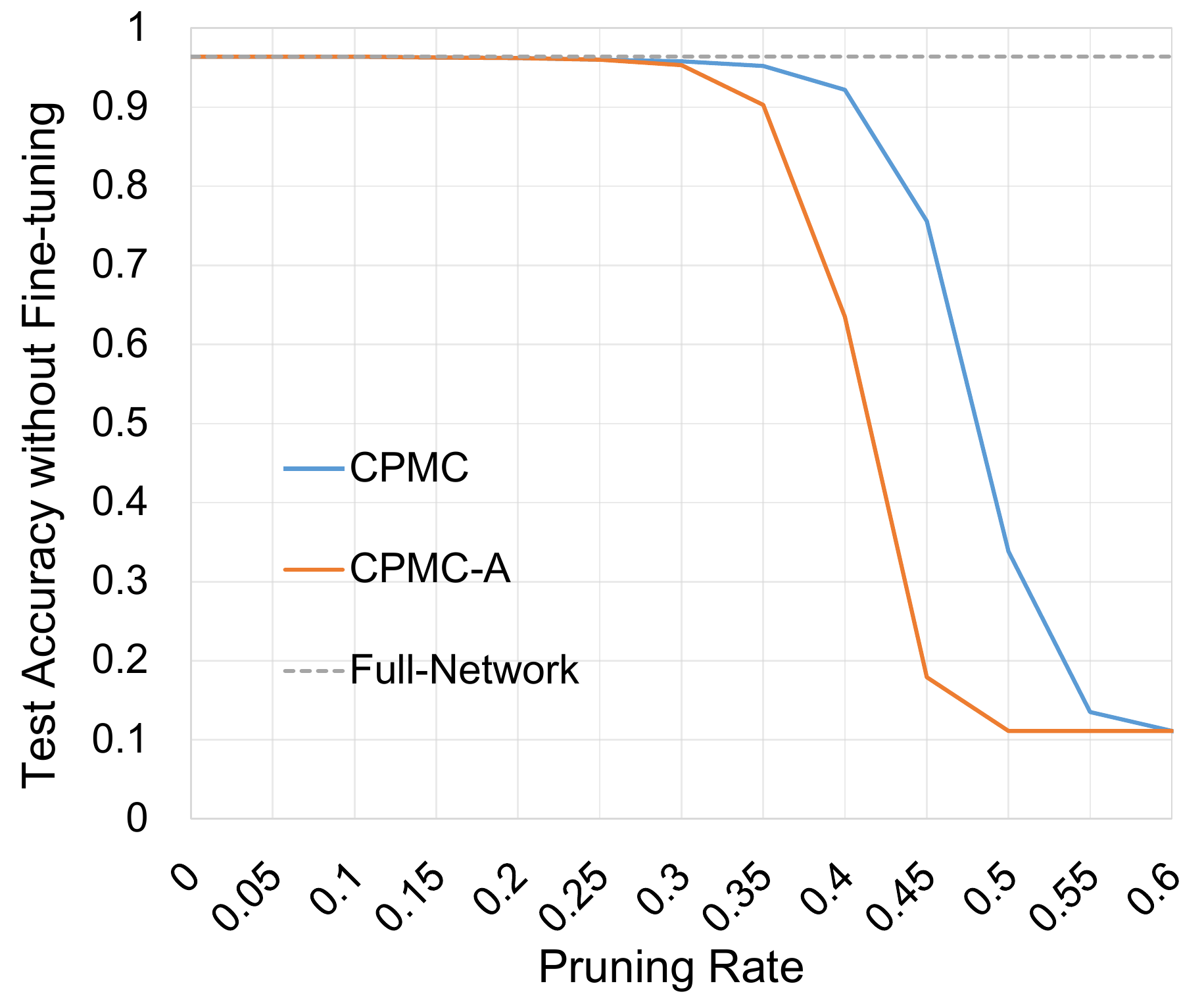}  
		\vspace{-5pt}
		\caption{Comparison between CPMC and CPMC-A which neglects weight dependency and only evaluates the out-channel of each channel for VGGNet on SVHN dataset. Obviously, CPMC has less accuracy drop compared with CPMC-A under the same prune rate.}
		\label{fig:line}
		\vspace{-2pt}
	\end{figure}

	\textbf{Effect of weight dependency}. We set a set of comparative experiments to show the effect of weight dependency. The result is shown in Figure \ref{fig:line}. CPMC-A is the method that prunes the channels only based on the evaluation of out-channel via multi-criteria. As the pruning rate of the model increases, the pruned model without fine-tuning by CPMC has less accuracy drop than CPMC-A under the same prune rate. The comparison result shows that CPMC is better than CPMC-A at identifying unimportant channels because CPMC takes weight dependency into consideration.
	
	\section{Conclusion and Future Work}
	\label{sec:CF}
	Current deep neural networks have many effects with high inference and storage costs. In this paper, we propose a novel and efficient channel pruning methods, namely CPMC, which have consideration of the weight dependency and use the multi-criteria to evaluate channels to compress more compact CNN models. Multi-criteria combines out-channel and in-channel to represent a channel and measures the importance of a channel in three aspects, including its associated weight value, the number of parameters, and computational cost. To avoid specifying the layerwise pruning ratios, we normalize the evaluation value to achieve cross-layer comparison. As a result, important channels and accuracy are greatly preserved by CPMC after pruning. Extensive experiments demonstrated the outperformance of CPMC to the other structural pruning algorithms. Furthermore, CPMC can also prune the light-weight CNN models such as ResNet-20, and get more compact models with little loss.

	\bibliographystyle{IEEEtran}
	\bibliography{IEEEabrv,main}

\end{document}